\documentclass[letterpaper, 10 pt, conference]{ieeeconf}  

\IEEEoverridecommandlockouts                              

\usepackage{graphicx}
\usepackage{multirow}
\usepackage{amssymb}
\usepackage{amsmath}
\usepackage{amstext}
\usepackage{comment}
\usepackage{enumerate}
\usepackage[dvipsnames]{xcolor}
\usepackage{pifont}
\usepackage{array}
\newcommand{\PreserveBackslash}[1]{\let\temp=\\#1\let\\=\temp}
\newcolumntype{C}[1]{>{\PreserveBackslash\centering}p{#1}}
\newcolumntype{R}[1]{>{\PreserveBackslash\raggedleft}p{#1}}
\newcolumntype{L}[1]{>{\PreserveBackslash\raggedright}p{#1}}
\usepackage[colorlinks=true,citecolor=blue,urlcolor=blue]{hyperref}
\usepackage{cite}
\usepackage{textcomp}

\title{\LARGE \bf
Integrating Artificial Intelligence and Augmented Reality in Robotic Surgery: An Initial dVRK Study Using a Surgical Education Scenario
}

\author{
Yonghao Long, Jianfeng Cao, Anton Deguet, Russell H. Taylor, and Qi Dou
\thanks{This project was supported by CUHK Shun Hing Institute of Advanced Engineering (project MMT-p5-20), Hong Kong RGC TRS Project No.T42-409/18-R, and InnoHK Multi-Scale Medical Robotics Center.}
\thanks{Y. Long, J. Cao and Q. Dou are with the Department of Computer Science and Engineering, The Chinese University of Hong Kong. 
A. Deguet and R. H. Taylor are with the Department of Computer Science, The Johns Hopkins University.
\textit{Corresponding author: Qi Dou (qidou@cuhk.edu.hk)}}%
}
\begin{document}
\maketitle
\thispagestyle{empty}
\pagestyle{empty}

\begin{abstract}
Robot-assisted surgery has become progressively more and more popular due to its clinical advantages. In the meanwhile, the artificial intelligence and augmented reality in robotic surgery are developing rapidly and receive lots of attention. However, current methods have not discussed the coherent integration of AI and AR in robotic surgery. In this paper, we develop a novel system by seamlessly merging artificial intelligence module and augmented reality visualization to automatically generate the surgical guidance for robotic surgery education. Specifically, we first leverage reinforcement leaning to learn from expert demonstration and then generate 3D guidance trajectory, providing prior context information of the surgical procedure. Along with other information such as text hint, the 3D trajectory is then overlaid in the stereo view of dVRK, where the user can perceive the 3D guidance and learn the procedure. The proposed system is evaluated through a preliminary experiment on surgical education task peg-transfer, which proves its feasibility and potential as the next generation of robot-assisted surgery education solution.
\end{abstract}

\section{INTRODUCTION}

Artificial intelligence (AI) and augmented reality (AR) are two important and increasingly essential techniques to be developed for next-generation robotic surgery. So far, AI and AR have individually focused on different perspectives. In specific, AI concentrates on recognizing and planning surgical activities in a way similar to what surgeons could do, based on computation and analysis of collected sensory data such as endoscopic videos~\cite{jin2021temporal, zhang2021swnet, 9336292} and robotic kinematics~\cite{van2020multi, su2018real, long2021relational}.
Recent advances of AI have substantially enhanced a number of tasks such as situation awareness of surgical procedures~\cite{garrow2021machine,huaulme2021micro} and automation of some actions with surgical robots~\cite{van2021gesture,huynhnguyen2021toward}.
In the meanwhile, AR aims to augment the surgical environment in a way to facilitate surgeon's operation and decision-making, based on visualization and integration of additional information that is computed offline or in real-time~\cite{volonte2013augmented,qian2019review}.
Equipped with an immersive view in surgical robotic console, AR has shown effectiveness for the education of novice surgeons~\cite{sheik2019next,barsom2016systematic,mcknight2020virtual} and is envisaged to be very helpful if could be adopted intra-operatively \cite{londei2015intra}. 


Unfortunately, to date, the advantages of AI and AR have not been merged in a sensible way for robotic surgery. 
The intriguing combination of AI and AR emerges as a versatile topic and has been exemplified in a number of application scenarios, such as games \cite{turan2019using}, driver training \cite{gabbard2019ar,boboc2020application} and virtual patients \cite{daher2020physical,gonzalez2020neurological}. Conferring intelligence on AR not only can boost the virtual experience but also harnesses the strong power of learning-based algorithms in demanding tasks such as surgical education. However, there have been few attempts in combining AI and AR for surgical robots. Some authors \cite{tanzi2021real,pan2020real,sadeghi2021virtual} have proposed to use computer vision models to localize the anatomy regions of interest, and then superimpose the results in the view of camera. Nonetheless, these solutions only account for presenting existing clues perceptively, without shedding lights on human-like decision behaviors. Concurrently, reinforcement learning (RL) is widely recognized as an effective method for skill learning~\cite{amini2020learning,kurach2019google,chiu2021bimanual,richter2019open}, but its potential is not fully exploited in the surgical robotics domain. One interesting scenario for exploring these issues is surgical education, in which the RL-based smart agent is expected to reason about surgical task and generate constructive guidelines for the novice. Such embodied intelligence is promising to significantly increase accessibility and reduce the cost of surgical training. Realisation of the intelligent guidance, in the form of AR visualization on surgical robotics platforms, could further enhance usability and user experience, yet how to achieve this goal remains unclear.

\begin{figure}[t]
    \centering
    \includegraphics[width=1.0\hsize]{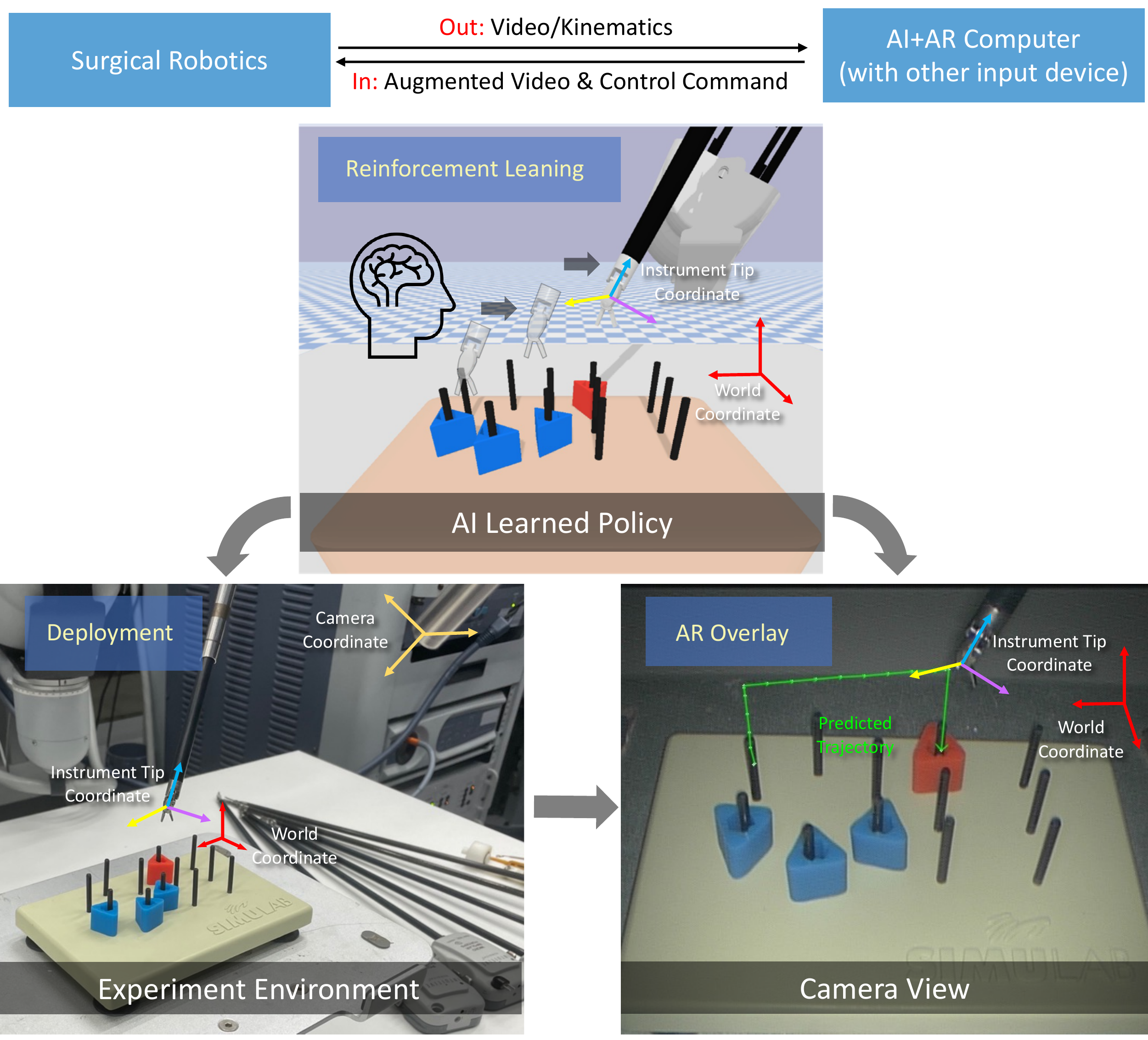}
    \caption{The integrated framework of AI and AR in robotic surgery on surgical education scenario of peg-transfer. The AI learns policy in a stimulated environment based on reinforcement learning and predict the guidance trajectory, which will be further visualized in the dVRK stereo viewer in the form of AR overlaid trajectory.}
    \label{TrainingFramework}
    \vspace{-0.50cm}
\end{figure}


In this paper, we aim to seamlessly integrate AI and AR, by augmenting RL-driven instrument movement trajectories with real-time AR visualizations, and implement the pipeline for a surgical education scenario using da Vinci Research Kit (dVRK), as shown in Fig.~\ref{TrainingFramework}. 
For the AI-enabled analysis, reinforcement learning is employed to learn a policy from the expert demonstration and through the interaction with the environment. Then the system can reason about the future action based on the current observation. 
This behavior can be embedded into an education process, where the automatically predicted actions could serve as helpful information to guide the trainee's movement step-by-step.
Subsequently, how to effectively visualize the information becomes very important, especially when incorporating with robotic platforms in real-time. 
We will demonstrate the feasibility of overlaying the 3D guidance trajectory within the stereo video via the dVRK console.
By projecting the 3D location of the trajectory generated from RL policy to the stereo video frames in the dVRK console, we can vividly observe the surgical scene with overlaid trajectory for the education purpose.
We have implemented and evaluated our method on the typical surgical education task of peg-transfer. 
To the best of our knowledge, this is the first work in exploring synergy of AI and AR with RL-based prediction and dVRK-based visualization for surgical education scenario. We hope to generate discussion and spark the potential of integrated benefits of AI and AR for surgical education and beyond.

\begin{figure*}[ht]
    \centering
    \includegraphics[width=0.79\hsize]{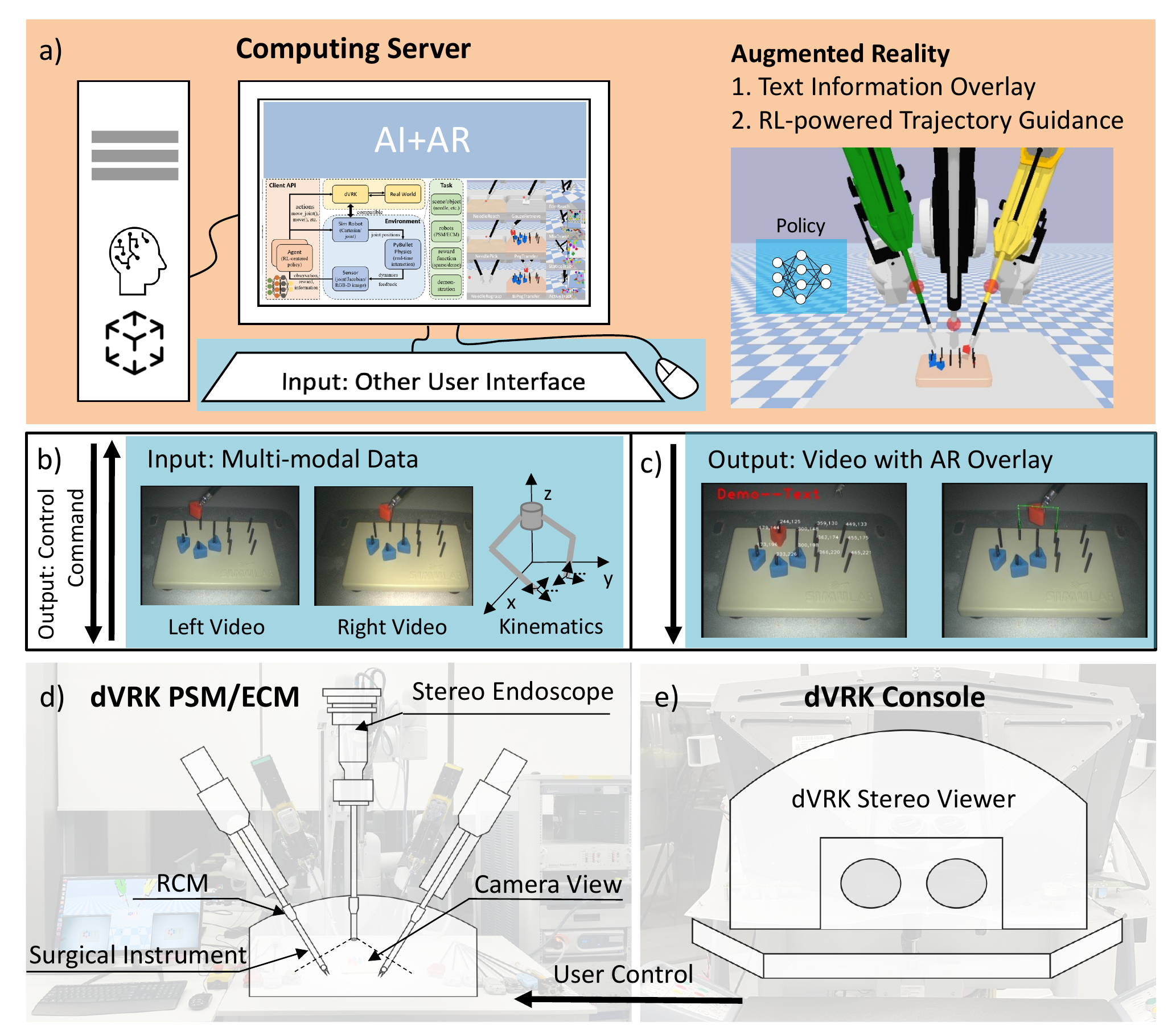}
    \caption{The pipeline of our proposed robot-assisted surgery education system based on dVRK. The computing server a) can acquire the stereo video and kinematics b) from the dVRK PSM/ECM d) and input the command to control the movement of dVRK PSM. With the AR and AI algorithms, the computer can overlay the AI and AR information on the video c) and then import the augmented video to the dVRK console e) for user to view.}
    \label{overall_Framework}
    \vspace{-0.50cm}
\end{figure*}

\section{RELATED WORK}

\subsection{Artificial Intelligence in Robotic Surgery}
The application of artificial intelligence in robotic surgery has been actively studied over the last decade~\cite{d2021accelerating} since the da Vinci Surgical System being clinically introduced in 2000~\cite{kazanzides2014open}.
Several topics have been found important and studied widely such as surgical instrument segmentation~\cite{zhao2020learning}, gesture recognition~\cite{long2021relational}, workflow recognition~\cite{jin2021temporal} and surgical scene reconstruction~\cite{long2021dssr}. They could support intra-operative decision \cite{blum2010modeling} and provide valuable database for surgical training and evaluation \cite{padoy2012statistical,nara2009surgical,neumuth2017surgical}. 
Although promising, these works just provide supplementary information without conceiving surgical plan, such as predicting the trajectories of the surgical instrument. Recently, the emergence of reinforcement learning opens the door to a new set of policy-based learning strategies \cite{rakhsha2020policy}. The efficacy of reinforcement learning has been revealed in surgical gesture classification \cite{liu2018deep,gao2020automatic}, surgical scene understanding \cite{torrents2020deep,chitsaz2009medical} and robot learning \cite{varier2020collaborative,chiu2021bimanual}. Through learning from expert demonstration, the RL agent could automatically generate meaningful solutions according to the task at hand. For example, in~\cite{chiu2021bimanual} and~\cite{richter2019open}, the authors propose to use Deep Deterministic Policy Gradients (DDPG) with Behavior Cloning (BC) to conduct surgical task of bimanual needles regrasping and autonomous blood suction. Both of them demonstrate encouraging results, showing that RL-based framework may potentially alleviate the demands on expert's guidance \cite{xie2020learning, loftus2020decision, nguyen2019new}.

Apart from the above, there have been some works using AI for surgical education, such as providing metrics and performance feedback based on the training records~\cite{sewell2008providing, nagyne2019robot, mirchi2020artificial} and differentiate expertise levels taking the stylistic characteristics into account \cite{ershad2018meaningful,liang2011surgical,ershad2019automatic}. But few of them consider its application to AR surgical education, where integrated AR and AI is highly demanded to guide the trainee vividly and automatically.
Given the superiority of RL introduced above, it is compelling to integrate RL into AR as an essential AI module, for example, a decision-maker \cite{wang2019lane, wierzbicki2000modern}, which encourages the AR system to generate content objectively.

\subsection{Augmented Reality for Robotic Surgery}
Augmented reality has been applied to robotic surgery in a variety of paradigms \cite{qian2019review}. 
In intraoperative applications, augmentations are superimposed in real-time to offer assistance: (i) enhance depth perception \cite{wen2017augmented,penza2017envisors,cutolo2017new}, (ii) compensate tactile sensory \cite{zevallos2018surgical}, (iii) expand field of view \cite{wang2018stereoscopic}, (iv) provide more intuitive human-machine interface \cite{wen2017augmented}, and (v) annotate helpful cues \cite{chen2017real,7150398}. 
Other applications utilize augmented reality for robotic surgery training \cite{tang2020review}. 
The common displaying media of augmented reality for robotic surgery include the da Vinci console, the computer monitor, and head-mounted display \cite{qian2019review}. For the purpose of surgical education, head-mounted display-based augmented reality is an advantageous medium as it enables 3D display and interactions for multiple users and the environment.
Jarc et al. apply augmented reality to a clinical-like training scenario, where 3D semitransparent tools controlled by the proctor are augmented and overlaid in the trainee's console as guidance \cite{jarc2017proctors}.
User study involving seven proctor-trainee pairs is conducted, where they demonstrate the augmented tools as an effective mentoring approach. However, currently most of the AR system have not taken AI as the core component for generating and creating the context-aware information.



\section{METHOD}

\subsection{Overall Framework}
The whole system is based on the dVRK \cite{dVRK} platform, the first standard and general da Vinci surgical system that has been open-sourced and further developed for other researchers to explore~\cite{d2021accelerating}. It has been widely used for the research on surgical imaging and perception~\cite{long2021relational}, control and hardware design~\cite{lin2019reliable}, system simulation~\cite{xu2021surrol} and surgical task automation~\cite{lu2020learning}, which substantiate the high reliability and flexibility of the platform. In this work, we propose to fully leverage the advantage of dVRK to integrate the augmented reality and artificial intelligent to form a robotic surgery education system as shown in Fig.~\ref{overall_Framework}. 
In total, our framework consists of several parts as described below:

\begin{enumerate}
\item AI+AR Computing Server, as shown in Fig.~\ref{overall_Framework} a): A high performance computer equipped with a high-end GPU for AR and AI algorithms deployment, which can acquire the stereo video and kinematics information from Patient Side Manipulator (PSM) and Endoscopic Camera Manipulator (ECM) on dVRK, receives input from other user interface, train and deploy RL policy or other AI framework and output the augmented video stream to monitor or other display devices.
\item Endoscope Video Acquisition and Kinematics Control: As depicted in Fig.~\ref{overall_Framework} b) and d). The dVRK contains a stereo endoscope on ECM which can capture the stereo video stream for 3D sensing and perception. In this case, we collect the video signal from the dVRK stereo endoscope with a video capture card to convert the video signal to the USB video stream, which can be retrieved by the computer. By using the Application Programming Interface (API) provided by dVRK Robot Operating System (ROS) packages, users can acquire the kinematics information (including tool tip position, velocity, rotation) from two PSMs on dVRK and input kinematics to control the movement of the PSMs. Also, the PSMs can be control directly by user using hands in the dVRK console.
\item Displaying AR video with dVRK console: As depicted in Fig.~\ref{overall_Framework} c) and e). Equipped with a stereo viewer in the dVRK console, users can view on the scene with overlaid AR and AI information from the computer in the surgical scenario. The stereo viewer can provide left and right views with disparity, where human can perceive the surgical video and overlaid information with 3D feelings. This function is later developed into TilePro$^\text{™}$~\cite{hyung2014tilepro} in the following generation of da vinci surgical robot, which showcases the potential possibility of applying on the real robotic surgery. 
\end{enumerate}

\subsection{Generating AI Guidance with Reinforcement Learning}
Instead of guiding the novice surgeon by experienced expert, our work proposes to generate guidance through advanced reinforcement learning automatically. Considering the intriguing superiority of SurRoL \cite{xu2021surrol} in leaning from demonstration and trials on dVRK, we attempt to incorporate it as our core AI module, learning and generating the guidance trajectory using the RL algorithm based on specific task.

SurRoL incorporates reinforcement learning to teach an intelligent agent to take interactive actions so that the cumulative reward can be maximized. While the agent takes action $a_t\in\mathcal{A}$ and interacts with the environment at state $s_t\in\mathcal{S}$, it receives reward $r_t\in\mathcal{R}$ based on whether the agent achieves the goal $g$ at step $t$.   From step $t=0$ to $t=T$, the trajectory can be formulated as $\tau=\{s_0, r_0, a_0, s_1, r_1, a_1...,s_T, r_T, a_T\}$. Besides, the translation $p(s_t|s_{t+1})$ is derived from the underlying physics engine. For robotic surgery, we define $a_t$ as six degrees of freedom, including position movement $(d_x, d_y, d_z)$ in Cartesian space, orientation $d_{\textit{yaw}}/d_\textit{pitch}$ in a top-down/vertical space setting and $j$ for the open ($j\geq 0$) or close ($j<0$) status of the jaw. Reinforcement learning targets on finding the policy $\pi$ to generate action $a_t=\pi(s_t)$. To promote the learning process, the reward $r_t$ is set to be goal-based, where the success function $f(s_t, g, a_t)$ would determine the reward $r_t$ by checking whether $a_t$ achieves the ground-truth status, such as 3D positions and 6D poses. Such scenario is similar to the attempting-feedback procedure in the surgical education. Finally, the target policy $\pi$ is learned by maximizing the experimental expectation $\mathbb{E}_\pi[\sum_{t=0}^T\gamma^tr_t]$, where $\gamma\in[0, 1)$ denotes the discount factor to balance the agent's attention on distant future and immediate future. 
In specific, we opt for a sample efficient learning algorithm called hindsight experience replay (HER)~\cite{andrychowicz2017hindsight} and combine it with Q-filtered behavior cloning. In practice, we will leverage a small amount of demonstration data generated from the scripted policies as the demonstration for imitation leaning. It may have great potential to be extensively developed and learn from large amount of surgery data and then generate the AI-powered guidance for surgical education.

\subsection{Real-time 3D Visualization by Augmented Reality}

To facilitate the visualization result and form a more intuitive education, we propose to use the AR techniques to design a 3D visualization system for surgical robots. 

\subsubsection{Different Coordinate System}
As an essential part of the AR system, the coordinates and the transformation among them serve as the basis of how we locate and align the AR information to camera view of the stereo endoscope. In our surgical education scenario, we denote the world coordinate as $\{W\}$, the coordinate of multiple objects in the experiment environment as $\{O_i\}$ with $i$ indicating the index of the objects, Remote Center of Motion (RCM shown in Fig.~\ref{overall_Framework} (b), which is a fixed base frame of the dVRK PSM during the operation) as $\{R\}$, the instrument tip as $\{P\}$ and the endoscope coordinate as $\{C\}$. And we use $T_{src}^{dst}$ to denote as the transformation from source $\{src\}$ to destination coordinate $\{dst\}$. During the deployment, $T_W^{R}$ are fixed and the same, and we manually place the objects in the same location as in the SurRoL. When given the kinematics information, we can compute the transformation of $T_R^P$ using the forward kinematics theory and control the PSM to move the instrument tip to the target pose $\{P\}$ in the SurRoL and the real environment as below.
If we have the transformation of the endoscope relative to the world $T_W^C$, we can calculate and get the position of the object relative to the endoscope $T_C^O$ using the following equation:
\begin{equation}
    T_C^{O} = T_W^{O} * T_C^{W};~~~~
    T_C^{W} = {T_W^{C}}^{-1}
    \label{equ_t}
\end{equation}

\subsubsection{Human-involved Flexible Calibration}
Calibration process is to acquire the coordinate of the endoscope relative to the world $T_W^C$. It is an very important step~\cite{tabb2017solving} for AR, because it can align the AR information with the environment. Some methods have been proposed to calibrate the hand-eye coordinate~\cite{tsai1989new, park1994robot}, but they needs at least an expert on robotic vision who use the calibration board to complete the whole process. Using the QR code to automatically retrieve the coordinate is another attempt and solution~\cite{qian2020ar}, but we need to carefully take the detecting condition into consideration and it may be restricted in the surgical scene. In our work, we propose an user-friendly method to easily calibrate the endoscope coordinate, which is more suitable for surgical education scenario. Specifically, we will first define some specific points and collect the 3D position $[X_W^i,Y_W^i,Z_W^i]$ of these points from SurRoL ($i$ denotes the point index). Then we manually locate the 2D location $[u_i,v_i]$ of theses points in the image. We design an User Interface (UI) as depicted in Fig.~\ref{Figure_calibration}, where users can easily locate and record the 2D location by clicking on the image using the mouse. According to the pinhole model of the camera, we can have the following equation:
\begin{equation}
\begin{bmatrix} u_i \\ v_i \\ 1 \end{bmatrix} = K * J * T_W^{C}\begin{bmatrix} X_W^i\\Y_W^i\\Z_W^i\\1 \end{bmatrix}
\label{equ_pnp}
\end{equation}
\begin{equation*}
K=\begin{bmatrix} f_x&0&c_x \\ 0&f_y&c_y\\ 0&0&1 \end{bmatrix};~~
J=\begin{bmatrix} 1&0&0&0 \\ 0&1&0&0\\ 0&0&1&0 \end{bmatrix}
\label{equ_pnp2}
\end{equation*}
where $J$ represents the perspective projection model and $K$ is the camera intrinsic matrix which can be straightforwardly acquired through~\cite{zhang2000flexible}. To solve and acquire the solution for $T_W^{C}$, we leverage the stable method called iterative solvepnp algorithm~\cite{lepetit2009epnp}. It is based on Levenberg-Marquardt optimization which minimizes re-projection error and finds the best solution. The algorithm needs only 3 points for getting the solutions (more points contributes to better solution), which is convenient and efficient for surgical application. 

\subsubsection{Augmented 3D Overlaying}
To overlay the 3D information on the image, we will first place the AR overlay information (3D) in the environment relative to the world coordinate $T_W^{\text{AR}}$, where we know the position $[X_W^\text{AR},Y_W^\text{AR},Z_W^\text{AR}]$. Then we will project the overlay information to the image using the projection equation same as the Eqn.~(\ref{equ_pnp}).
Specifically, we sometimes may need different views for AR overlaying. And we can use the proposed calibration method to calibrate the coordinate and overlay the new generated AR information relative to the new novel view.
This is very promising as in the surgical education scenario, multiple views may enrich the information and help the surgeon to learn better.

\subsection{Integrating AI and AR for Trajectory Guidance}\label{Sec:Deployment}

With the AI and AR components introduced above, we further propose to integrate these two modules to form a AI-powered trajectory guidance with AR visualization. 
Firstly, the dVRK stereo endoscope will capture the stereo video from the surgical scenario and output for computing server. Then we will use the proposed calibration method to calculate and align the coordinate of the endoscope and world. We will then train the RL policy to learn on the specific tasks in the SurRoL. Given a well-learned policy and providing with the known state of the environment or the observation, we can generate the predicted trajectory from the action, which is represented in the coordinate of the world. Afterwards, we can project the trajectory points to the video frames using the projection equation. By lining up the points, we can finally generate a trajectory guidance. With the augmented video displayed in the stereo viewer of dVRK console, the user can perceive the 3D AR trajectory guidance for the purpose of robotic surgery education.

\section{Experiment}
\begin{figure}[t]
\vspace{+2mm}
    \centering
    \includegraphics[width=0.9\hsize]{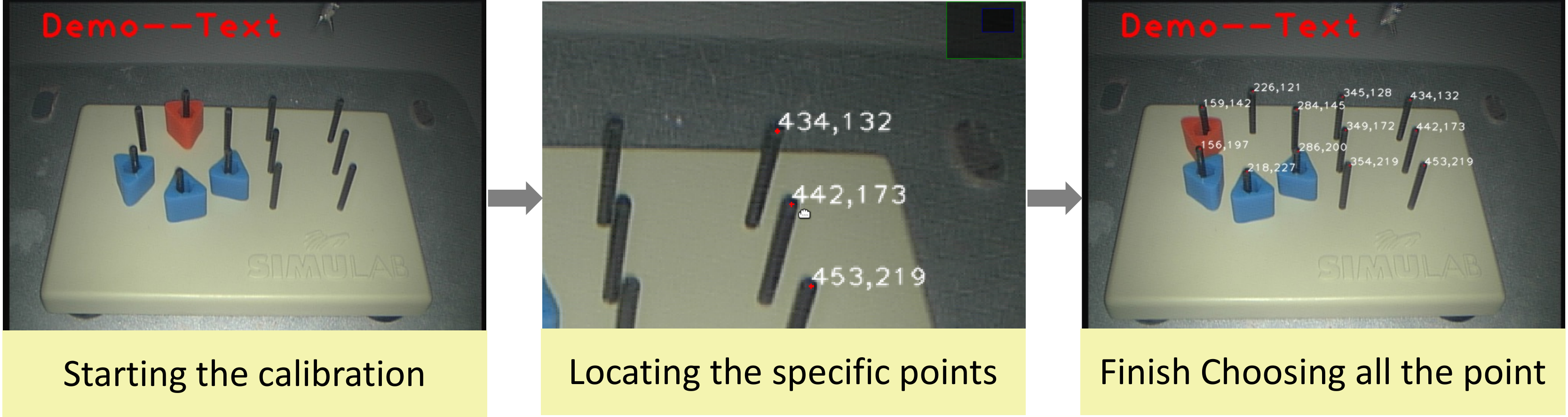}
    \caption{Our proposed human-involved calibration process where we can flexibly locate the specific points on the peg board for calibrating.}
    \label{Figure_calibration}
\end{figure}
\begin{figure}[t]
    \centering
    \includegraphics[width=1.0\hsize]{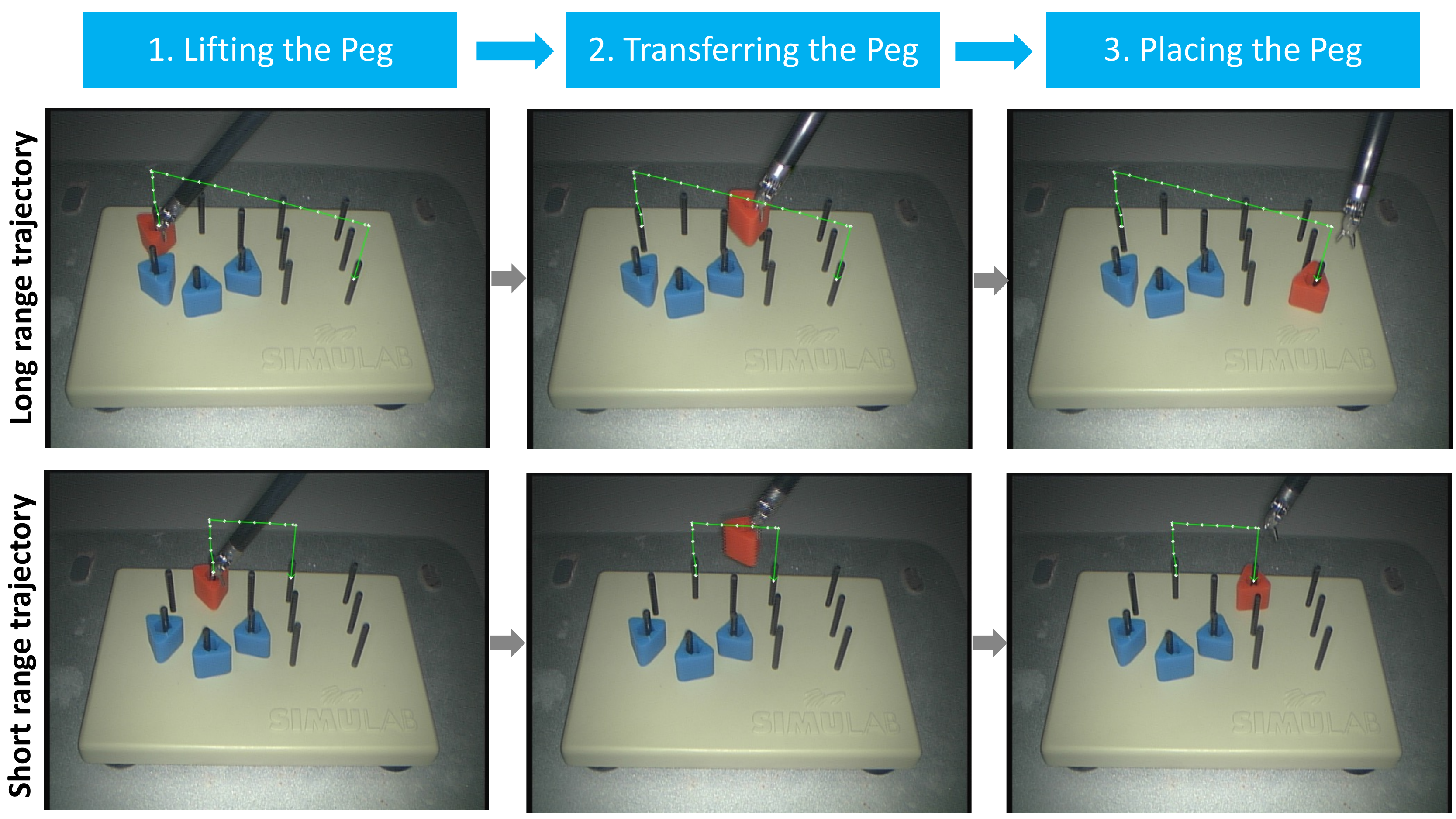}
    \caption{Two trajectories overlay visualization results with long range transfer (top) and the short range transfer (bottom).}
    \label{Figure_exp}
    \vspace{-0.4cm}
\end{figure}

To study the preliminary results of the proposed method, we choose the peg-transfer task as the experimental task in our work, which is one of the Fundamentals
of Laparoscopic Surgery (FLS) tasks for surgical skills education~\cite{joseph2010chopstick}. In this experiment, we move the block from one peg to the other peg with one PSM on the dVRK. The whole process contains three procedures: 1) lifting the peg, 2) transferring the peg and 3) placing the peg, as depicted in Fig.~\ref{Figure_exp}. We train the RL policy on this task and generate the block transferring trajectory based on the trained model. Then we overlay the peg-transfer path as the AR trajectory on the stereo viewer for visualization.

We conduct the experiment on a computer using ubuntu 18.04 as the system,  which is equipped with 8 cores 3.60GHz Intel$^\text{®}$ Xeon(R) W-2123 CPU, NVIDIA$^\text{®}$ TITAN RTX GPU and 16G RAM. All the algorithms are written in Python. The image resolution from the endoscope we use is $640\times480$.

\subsection{Reinforcement Learning on Peg-transfer}

In this work, we follow the peg-transfer setting in SurRoL\cite{xu2021surrol} where the goal tolerate distance is set as 0.5 cm in a workspace with size of 10 $\text{cm}^2$. Each episode lasts for 50 timesteps. When the episode ends, PSM reaches out of the workspace or the peg-transfer is completed, the environment will be reset and the positions of initial peg and target peg will be resampled. Two evaluation settings are proposed, (1) short range: the block moved by user will not encounter any obstacle peg during the transfer; (2) long range: the block needs to avoid 1 to 3 pegs to finally reach the target peg, as these two situations are similar to simple and complicated surgical scenario for education. We randomly generate 100 scripted demonstrations for RL to imitate and train for 100 epochs in the training stage. In the testing stage, for each experiment, we conduct the peg transfer for 1000 trials (with 500 short range and 500 long range trials) and evaluate the success rate. We select top 3 performed RL policy and average their results. The results are shown in the Table~\ref{table:RL}, where we can find that the RL policy can perform well on long range while achieving a higher performance on the short range peg-transfer with success rate 92.6\%. Overall, our RL can achieve promising result with average success rate 86.4\% on peg-transfer.

At the deployment stage, we export the complete predicted trajectory from the RL policy and input the trajectory location to control the PSM of real dVRK to conduct peg-transfer. We show a failure case from the long range setting in Fig.~\ref{Figure_fail} a), where the block is not placed to the target peg properly because of the biased trajectory generated from the RL policy. Although there exists failure trial, the above results already bring out a lot of possibilities of applying more advanced RL algorithms on the complex robotic surgery education scenario. 


\begin{table}[t]
\vspace{+2mm}
    \caption{
    \textbf{The evaluation of \textit{Peg-Transfer} on SurRoL.}
    }
    \vspace{-6pt}
    \centering
    \begin{tabular}{c|cc}
    \hline
    Distance       & Trials & Success Rate (\%) \\ \hline
    Short range    & 463/500  &  92.6                \\
    Long range     & 401/500  &  80.2    \\ \hline
    All            & 864/1000  &  86.4
    \\ \hline
    \end{tabular}
    \label{table:RL}
    \vspace{-0.4cm}
\end{table}

\subsection{Visualization and Analysis on the AR results}
In the calibration process, we choose the top center points of the twelves pegs as the specific points $[u_i,v_i]$ for solving the coordinate of the endoscope. As we can see in Fig.~\ref{Figure_calibration}, the specific points are located and shown in red dot. The point locations are shown next to these points with white text. After pointing out 12 points on the peg-transfer board, we can finish calibrating the coordinate. 

As shown in Fig.~\ref{Figure_exp}, we show two cases from our experiments, with the top one showing the peg-transfer with long range distance and the bottom one showing the short range distance. In the experiment, we deploy our RL policy to generate the 3D trajectory and project the trajectory to the video frames. As we can see in the figure, the white dots in the image frame represent each action step in the trajectory, and the green lines represent the trajectory. It can be observed that our method can intuitively and accurately overlay the trajectory in the video frames, where the instrument tip will follow the overlaid trajectory visually in the experiment.

\begin{figure}[t]
    \centering
    \includegraphics[width=0.8\hsize]{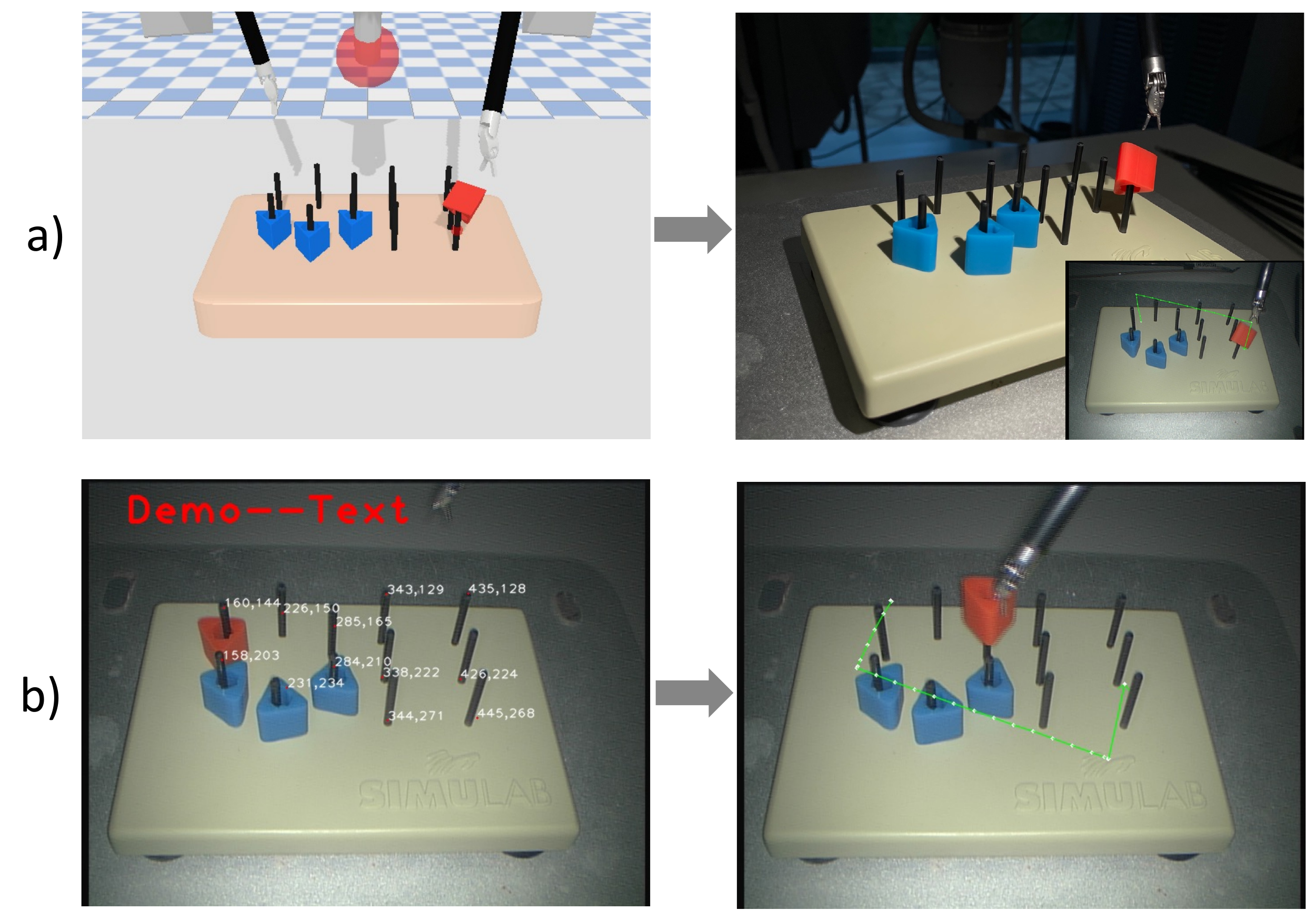}
    \caption{The failure cases with (a) because of the biased
 trajectory generated from RL and (b) due to manually caused calibration error.}
    \label{Figure_fail}
    \vspace{-0.5cm}
\end{figure}

To further analyse the stability and precision of the calibration, we randomly find 5 new users with engineering background to conduct the calibration process. We calculate the re-projection error (pixel) \cite{koide2019general} based on twelves specific points on the pegs. Each user conducts 10 trials and we report the mean and standard deviation of the error in Table~\ref{table:error}. From the table we can find that average error is within 1 pixel (0.76) which is precise for the visualization purpose. Besides, we only observe one failure trial where the trajectory is calculated and projected to the wrong location due to the large error from manually locating the specific points, as shown in Fig.~\ref{Figure_fail} b). The above results demonstrate the well performance of the calibration process and AR visualization for trajectory guidance on peg-transfer education scenario.

\begin{table}[tp]
\vspace{+2mm}
    \caption{
    \textbf{The evaluation of AR calibration on \textit{Peg-Transfer} scenario.}
    }
    \centering
    \vspace{-1mm}
     \resizebox{.95\columnwidth}{!}{
    \begin{tabular}{c|ccccc}
    \hline
    Re-project error (pixel) & User1 & User2 & User3 & User4 & User5 \\ \hline
    Average value       & 0.77  & 1.02  & 0.70 &  0.61  & 0.71           \\
    Standard deviation  & 0.20  & 0.14  & 0.12 &  0.10  & 0.15            \\ \hline
    \end{tabular}}
    \label{table:error}
\end{table}

\subsection{System Latency Study}
As a systematic assessment of the whole pipeline, latency is more than important for a reliable and real-time system. In our work, we mainly focus on studying two different types of latency that may be introduced by the system. One is the video capturing and displaying latency, which is often caused by the delay of the video capture device, video signal transfer and the monitor. The other is overlaying latency, which is mainly caused by computing and projecting the points from 3D location to 2D image, as well as the time for drawing the information on the video frame.

\begin{figure}[t]
    \centering
    \includegraphics[width=1\hsize]{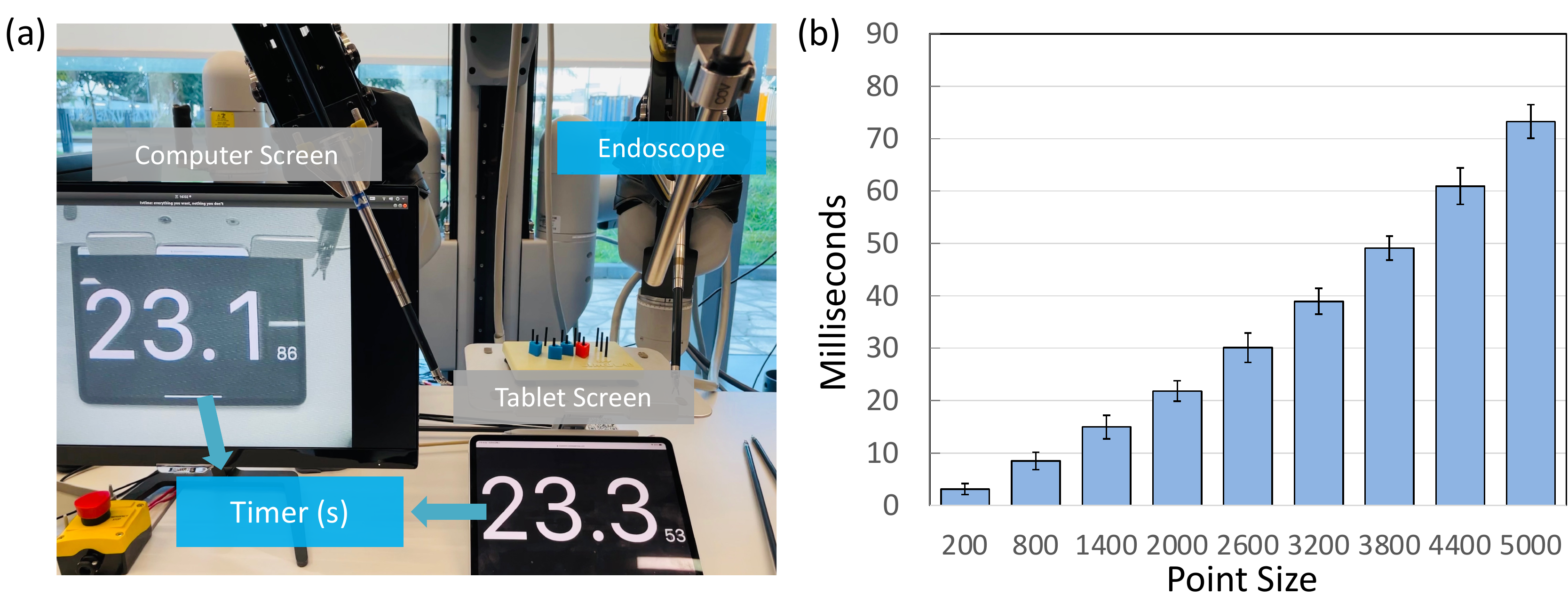}
    \vspace{-18pt}
    \caption{(a) Illustration for evaluating video capturing and displaying latency and (b) chart showing the latency related to the point size.}
    \label{Figure_latency}
    \vspace{-0.5cm}
\end{figure}

To evaluate the video capturing and displaying delay, we place a tablet under the camera view of the endoscope which displays the timestamp on the screen. Then we capture the video from endoscope camera and display it on the screen of the computer. Afterwards, we take a photo using another camera, which will include both table screen and computer screen. The difference of two timestamps shown on two screens is the latency. The illustration can be seen in Fig.~\ref{Figure_latency} (a). Finally, we find the average latency is around 161 milliseconds with standard deviation of 17 milliseconds from 100 samples. Which need to be mentioned is that the latency may vary when given different video capture equipment or monitor. According to some studies~\cite{miller1968response, tolia2006quantifying} and our own observation among all the trials, the latency is delicate enough that the human would not be able to perceive and will not significantly affect the operating.

To further evaluate the overlaying latency, we conduct an experiment which measures the computing time of overlaying (projecting 3D trajectory points to the video frames and visualize them).  We evaluate the time cost when given different size of trajectory points (200-5000), as shown in Fig.~\ref{Figure_latency} (b), where we randomly generate the 3D locations of these points and repeat the experiment for 300 times. Finally we report the average and standard deviation of the time cost. From the figure we can observe that when given more points, the computation time may increase showing exponential growth. When around or less than 2600 points, we can achieve the real-time process ($>30$ Hz). In our peg transfer scenario, the RL will generate around 100 trajectory points which cost just about 1 ms for projecting and overlaying, it is more than enough for a real-time visualization. According to the above results, our designed system can efficiently overlay the trajectory on video frames forming a real-time AR visualization, which has great potential for more complicated real-time AR robotic surgery education scenario.

\section{CONCLUSIONS AND FUTURE WORK}
In this work, we confer the intelligence on AR by integrating AI module as a decision-maker for generating trajectory. The AI module with reinforcement learning is leveraged to learn policy-based operation and dynamically generate the trajectory based on user's task. To provide an immersive visualization and facilitate intuitive education, the trajectory plan is further projected and overlaid onto the stereo video, assisting the trainee to perceive the intelligent guidelines in 3D AR. A simple yet effective calibration with human-computer interaction is also designed to realize coordinate calibration through an user interface. We conduct the experiment with the surgical training on task peg-transfer, which demonstrates the feasibility and potential of AI-powered AR for robotic surgery education. 

In our future work, we shall further exploit how to design and incorporate more AI and AR modules into the system, such as virtual tool guidance, overlaying the result of surgical workflow recognition and instrument segmentation. Furthermore, we shall integrate network communication for realizing a remote surgery education. Methods of measuring the operation of surgeon compared to the generated 'expert' trajectory and user study may also be included for evaluating the effectiveness of the proposed system.

\newpage
\bibliographystyle{IEEEtran}
\bibliography{IEEEabrv,ref}

\end{document}